\documentclass[runningheads]{llncs}
\usepackage[T1]{fontenc}
\usepackage{graphicx}
\usepackage{tikz}
\usetikzlibrary{shapes.geometric, arrows, positioning, calc, shapes.multipart}
\usepackage{forest}
\usepackage{tabularx}
\usepackage{array}
\usepackage{multirow}
\usepackage[utf8]{inputenc}
\usepackage{graphicx}
\usepackage{multirow}
\usepackage{makecell}
\usepackage{array}
\usepackage{amsmath}
\usepackage{subcaption}
\usepackage{xcolor}

\definecolor{color1}{HTML}{fee3ce}
\definecolor{color2}{HTML}{d2e2ef}
\definecolor{color3}{HTML}{D9EADF}

\definecolor{color1d}{HTML}{b7282e}
\definecolor{color2d}{HTML}{313772}
\definecolor{color3d}{HTML}{376439}

\begin{document}

\title{A Survey on Enhancing Causal Reasoning Ability of Large Language Models}

\author{Xin Li \textsuperscript{*} 
        \and Zhuo Cai \textsuperscript{*}
        \and Shoujin Wang \textsuperscript{\dag}
        \and Kun Yu\and Fang Chen}
\authorrunning{X. Li et al.}
\institute{University of Technology Sydney, Ultimo NSW 2007, Australia}
\maketitle

\renewcommand{\thefootnote}{}
\footnotetext{\textsuperscript{*} Equal contribution.\\
\textsuperscript{\dag} Corresponding author.}

\begin{abstract}
Large language models (LLMs) have recently shown remarkable performance in language tasks and beyond. However, due to their limited inherent causal reasoning ability, LLMs still face challenges in handling tasks that require robust causal reasoning ability, such as healthcare and economical analysis. As a result, a growing body of research has focused on enhancing the causal reasoning ability of LLMs. Despite the booming research, there lacks of a survey to well review the challenges, progress and future directions in this area. To bridge this significant gap, we systematically review literature on how to strengthen LLMs’ causal reasoning ability in this paper. We start from the introduction of background and motivations of this topic, followed by the summarization of key challenges in this area. Thereafter, we propose a novel taxonomy to systematically categorize existing methods, together with detailed comparisons within and between classes of methods. Furthermore, we summarize existing benchmarks and evaluation metrics for assessing LLMs’ causal reasoning ability. Finally, we outline future research directions for this emerging field, offering insights and inspiration to researchers and practitioners in the area.



\end{abstract}

\section{Introduction}
Large language models (LLMs) have garnered significant attention for their exceptional performance across diverse natural language processing (NLP) tasks and beyond \cite{achiam2023gpt,dubey2024llama}. Leveraging their powerful real-world knowledge and generalization capabilities, LLMs have spearheaded innovation in various practical domains, including healthcare \cite{gopalakrishnan2024causality}, finance \cite{wu2023bloomberggpt}, psychology \cite{tong2024automating}, etc.

Although LLMs have achieved remarkable success, relying solely on their inherent reasoning ability is insufficient to effectively address complex causal tasks \cite{jin2023can,kiciman2023causal}. This limitation arises because LLMs lack genuine causal reasoning ability \cite{mirzadeh2024gsm}. Instead, they often replicate the reasoning steps found in the training data. When the training data contain incomplete or incorrect causal relationships, LLMs struggle to effectively handle causal reasoning tasks.

To address this issue, various approaches have been proposed in recent years to enhance the causal reasoning ability of LLMs, enabling them to better adapt to real-world scenarios that demand robust causal reasoning ability \cite{pais2024large,chiunveiling,jin2023cladder,pan2024unifying}. Most of these works have been published in top conferences and journals, such as Nature Medicine, NeurIPS, TKDE, MM and others. This highlights the significant research values of this topic, and researchers' substantial attention to it. However, there lacks of a survey to well review the challenges, progress, and future directions in this area, hindering further development in this area. To bridge this gap, in this paper, we systematically review existing methods aimed at enhancing the causal reasoning ability of LLMs. To deepen understanding in this research area, this section outlines the key motivations driving its development, compares with related survey papers, and highlights our contributions.

\subsection{Motivations: Why enhance LLMs’ causal reasoning ability?}

\textbf{LLMs' inherent causal reasoning ability is limited.}
Applications of LLMs have demonstrated remarkable capabilities, especially Chatbots, which include the ChatGPT series \cite{achiam2023gpt}, Llama series \cite{dubey2024llama}, etc. However, their inherent limitations in causal reasoning remain a significant challenge, as evidenced by their under-performance in tasks requiring a deep understanding of causality \cite{mirzadeh2024gsm}. The core mechanism of LLMs relies on generating responses based on pre-trained data and user-input tokens, leveraging the Transformer architecture \cite{huang2023advancing}. This architecture predicts the probability of the next token by analyzing contextual relationships, primarily through calculating inter-token distance. While this process effectively captures correlations within the data, it does not inherently distinguish between correlation and causation~\cite{wu2024causality}. This limitation becomes particularly pronounced in domain-specific datasets, where spurious correlations can mislead the LLMs and result in inaccurate or unreliable outputs.\\

\noindent \textbf{LLMs are black-box models which can cause unknown risks for downstream applications.} LLMs’ internal mechanisms are black-box based and are not transparent, and thus they have a high possibility of causing unknown risks for downstream applications \cite{zhao2024explainability,manakul2023selfcheckgpt}. For instance, the black-box nature of LLMs makes it difficult to trace and interpret the logical chain behind their generated content. Such lack of explainability can undermine the trustworthiness of LLMs, where non-transparent outputs may lead to severe consequences. Additionally, the black-box property can result in unpredictable outputs, causing LLMs to generate inappropriate or even harmful content (e.g., misinformation) in certain situations. Enhancing the causal reasoning ability of LLMs can improve the interpretability and transparency of their generated outcomes. By tracing the underlying causal relationship network, it becomes easier to detect the causes of harmful or non-factual outcomes and correct them promptly. \\

\noindent \textbf{Real-world scenarios require high causal reasoning ability.}
Due to the widespread application of LLMs across multiple domains, they are no longer limited to simple question-answering. Currently, LLMs are now also applied to solve increasingly complex reasoning tasks, making it essential to enhance their reasoning capabilities. For instance, by integrating LLMs and specific prompting strategies, LLMs’ performance has significantly increased when processing reasoning tasks such as multi-step logical reasoning problems\cite{creswell2022selection,liu2023concise}. However, they can not perform optimally if they only rely on their inherent reasoning ability. \\

\noindent \textbf{Enhancing LLMs' causal reasoning ability can improve their decision-making performance.}
LLMs exhibit limited decision-making ability in many reasoning tasks like \cite{huang2024far}, partly due to the incompleteness of their pre-trained datasets, which often lack critical causal relationships between variables \cite{eigner2024determinants}. Addressing these gaps in causality within LLMs can not only enhance their causal reasoning but also strengthen their decision-making performance \cite{wang2024causal}. Since decision-making in LLMs often involves predicting the potential outcomes of various interventions, robust causal reasoning enables them to simulate the consequences of specific interventions, leading to more optimal decisions.

\subsection{Comparison with related survey papers and our contributions}
Several prior surveys have explored topics related to combining LLMs and causality. For instance, an existing survey focuses specifically on the different types of reasoning in LLMs and strategies to improve their performance in these contexts, encompassing deductive reasoning, inductive reasoning, abductive reasoning, as well as formal and informal reasoning \cite{huang2022towards}. Additionally, some studies emphasize the integration of LLMs with causal inference \cite{liu2024large} and causal discovery \cite{wan2024bridging}. Moreover, this survey narrows its scope to the application of LLMs in addressing various causal tasks and evaluating their intrinsic causal reasoning capabilities \cite{ma2024causal}. Another work examines how LLMs can enhance causal reasoning performance, focusing on their diverse roles in reasoning tasks, whether as reasoning engines or as complementary tools supporting traditional methods \cite{xiong2024improving}. 

In contrast, our survey adopts a novel perspective by comprehensively investigating methods aimed at enhancing the intrinsic reasoning capabilities of LLMs. Unlike other surveys mentioned above, which focus on integrating LLMs with external causal reasoning techniques or applying LLMs to solve causal reasoning tasks, our survey has a distinct starting point and scope. Additionally, our survey provides a novel taxonomy that categorizes existing methods from the perspective of whether they are domain knowledge driven or model driven.

\subsubsection{Contributions.}
The main contributions of this paper are summarized as:
\begin{itemize}
    \item We systematically review existing methods for a burgeoning and cutting-edge research area: enhancing the causal reasoning ability of LLMs.
    \item We outline some key challenges in the process of enhancing LLMs' causal reasoning ability, providing insights for better understanding of this topic.
    \item We provide a comprehensive summary of existing methods, categorized as either domain knowledge driven or model driven approaches. 
    \item We offer a detailed comparison of different classes of methods, outlining their respective strengths and weaknesses to guide researchers in selecting methods suited to specific tasks.
    \item We summarize the commonly used benchmarks and evaluation metrics for assessing the causal reasoning ability of LLMs, aiming to assist researchers in advancing their progress in this field.
    \item We provide valuable future prospects in terms of LLMs' causal reasoning. These future directions can provide researchers in this area with more insightful guidance on how to further advancing this domain.
\end{itemize}

\section{Challenges in LLMs’ Causal Reasoning Process}
While enhancing the causal reasoning ability of LLMs is of great significance, there are multiple challenges in this process. We have identified three potential challenges: (1) Difficulty in constructing appropriate prompts to guide LLMs' causal reasoning; (2) LLMs' outdated knowledge for causal reasoning; and (3) LLMs' causal hallucination issues.

\subsection{Difficulty in Constructing Appropriate Prompts to Guide LLMs' Causal Reasoning}
With the development of LLMs, prompt engineering has emerged as an important field. Prompts are natural language instructions that provide context to guide LLMs and activate LLMs' relevant internal knowledge in response to queries, with the goal of generating ideal results \cite{sahoo2024systematic}. However, designing effective prompts to activate and enhance LLMs' causal reasoning ability is a potential challenge. Even when utilizing the same prompt to guide an LLM through identical causal reasoning tasks, the outcomes can still be different. This inconsistency causes significant difficulties for tackling causal reasoning tasks, it is hard to achieve ideal results for LLMs.

Currently various efforts have focused on designing effective prompt templates to enhance LLMs' performance in different causal reasoning scenarios. Existing classic prompt design methods include Chain-of-Thought (CoT), Program-of-Thoughts (PoT), and ReAct-style Prompting \cite{liu2024llms}. The CoT prompting method is aiming to tackle complex NLP tasks by incorporating intermediate reasoning steps \cite{liu2024llms}. Program-of-Thoughts (PoT) is a kind of methods to guide LLMs to generate code programs based on solutions for reasoning tasks \cite{liu2024llms}. In contrast, ReAct-style prompting is a reasoning agent, which integrate reasoning and action in LLMs to solve reasoning tasks. Additionally, the Causal Chain of Prompting (C2P) is developed as a new novel prompting framework, and presents the significant performance across various real-world causal reasoning datasets \cite{bagheri2024c2p}.

\subsection{LLMs' Outdated Knowledge for Causal Reasoning}
The training of LLMs relies on the large amounts of datasets. As the novel research, new policies, and real-time events appeared, outdated knowledge within LLMs can cause the disadvantages about accuracy and relevance of their generated outputs, especially in rapidly evolving fields. Additionally, the lack of specific domain and real-time factual knowledge limits LLMs’ utility in certain contexts \cite{wang2023survey}. For instance, LLMs' generated outcomes may lag behind the real-world environment if LLMs' training datasets do not include real time information. Therefore, solving issues that caused by LLMs' internal knowledge, can be effectively pathway to enhance LLMs' causal reasoning ability.

Retrieval-augmented generation (RAG) is a method designed to solve this challenge by using retrieval techniques to fetch relevant information from external sources, thereby supporting real-time answer generation \cite{xu2024knowledge}. Additionally, knowledge editing is employed to directly update the parametric knowledge of LLMs, keeping them align with real-world information \cite{de2021editing,xu2024knowledge}.

\subsection{LLMs’ Causal Hallucination Issues}
With the development of LLMs, they have demonstrated strong performance in a variety of NLP tasks. However, LLMs are also prone to hallucination issues, where generated outcomes according to prompts seem reasonable but is actually inconsistent with real-world facts. \cite{huang2023survey}. This can decrease LLMs' performance for causal reasoning tasks. Therefore, addressing these hallucination issues and ensuring that generated outcomes are based on accurate knowledge aligned with inputting prompts remains a key challenge in enhancing LLMs' reasoning ability.

To address this challenge, integrating knowledge graph (KG) into LLMs' reasoning process has proven to be a feasible approach to solve complex causal reasoning tasks \cite{pan2024unifying}. On the other hand, fine-tuning methods for LLMs can make the contribution to mitigate the hallucination issues in these tasks. For example, lightweight fine-tuning methodologies such as freeze tuning, adapter tuning and prompt tuning (P*-tuning), present the significant performance in experiments and are effective to enhance LLMs' causal reasoning ability \cite{10366620}.

\section{Progress on Enhancing LLMs' Causal Reasoning Ability}
\begin{figure}[ht]
\centering
    \begin{forest}
        for tree={grow=east, parent anchor=east, child anchor=west, draw, rounded corners,  align=center, 
        minimum width=3.5cm,
        inner sep=4pt,
        edge path={\noexpand\path [draw, \forestoption{edge}] (!u.parent anchor) -- ++(8pt,0) |- (.child anchor)\forestoption{edge label};}
        }
        [Enhancing LLMs’\\Causal Reasoning Ability, draw=color1d, fill=color1
            [Model Driven \\ Methods, draw=color2d, fill=color2
                [Counterfactual\\Reasoning, draw=color3d, fill=color3]
                [Causal Effect\\Estimation, draw=color3d, fill=color3]
                [Causal Graph\\Construction, draw=color3d, fill=color3]
            ]
            [Domain Knowledge \\Driven Methods, draw=color2d, fill=color2
                [Fine-tuning, draw=color3d, fill=color3]
                [Pre-defined Prompt, draw=color3d, fill=color3]
                [Contextual \\Knowledge, draw=color3d, fill=color3]
                [Domain Experts, draw=color3d, fill=color3]
            ]
        ]
    \end{forest}
\caption{Classification of methods for enhancing LLMs' causal reasoning ability}
\label{fig:Classification1}
\end{figure}
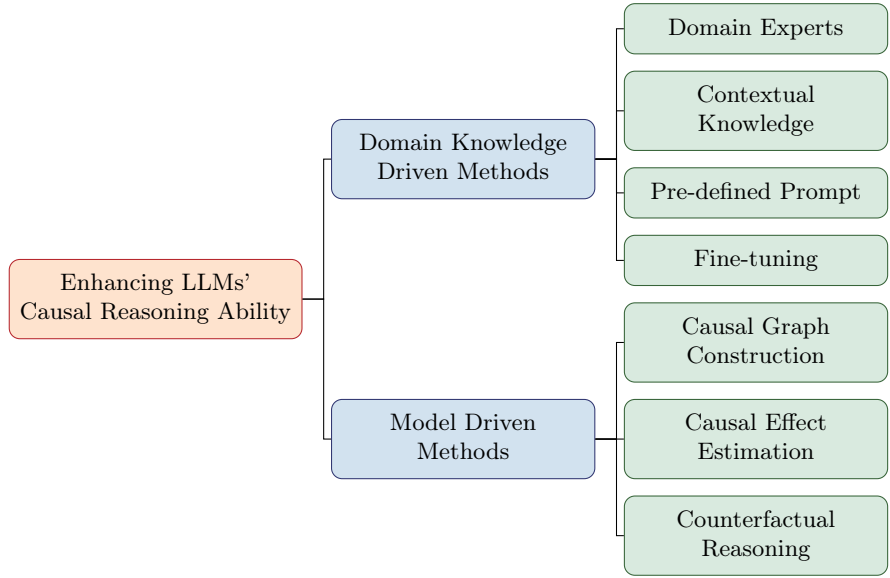

\begin{figure}[t]
    \centering
    \includegraphics[width=1\linewidth]{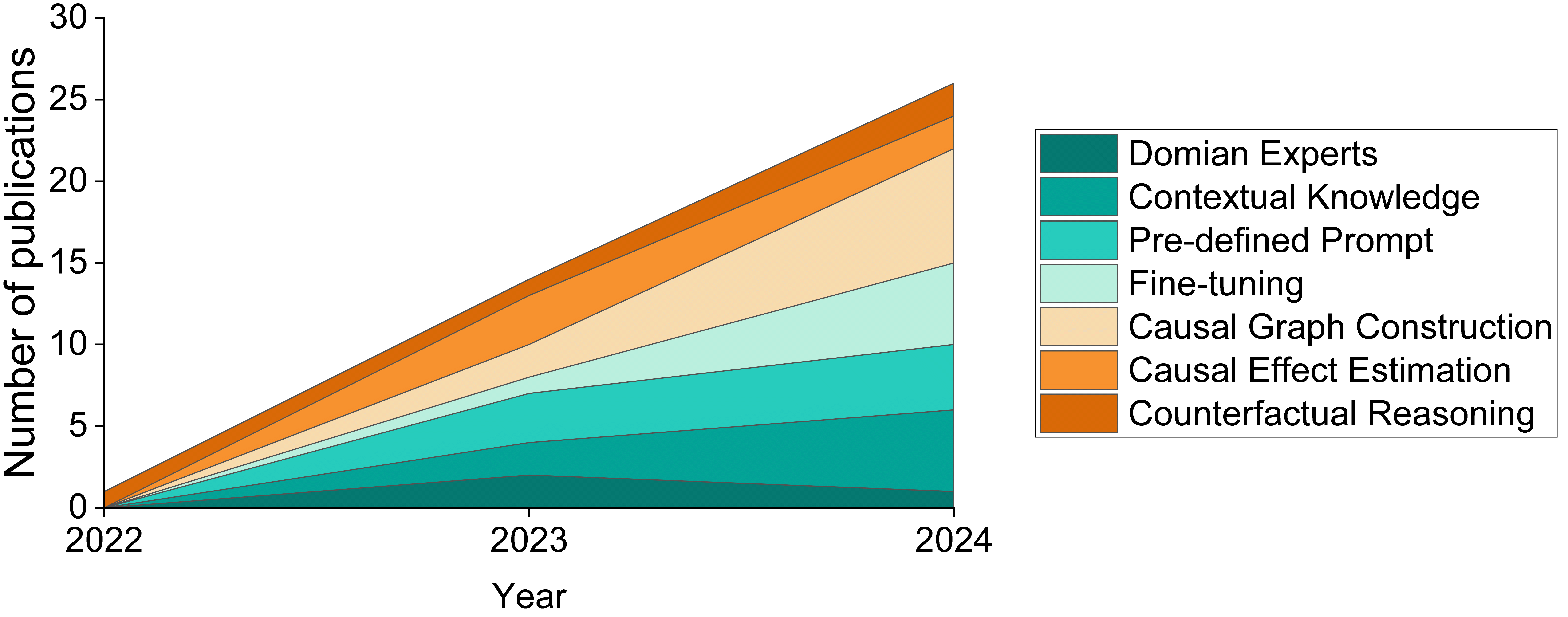}
    \caption{Number of publications in each class per year}
    \label{fig:statistic_per_year}
    \vspace{-1em}
\end{figure}

We categorize existing methods on this topic from a novel perspective: whether they are domain knowledge driven or model driven. These two categories are further divided into several subcategories, as illustrated in Fig.\ref{fig:Classification1}. Additionally, Fig.\ref{fig:Classification1} provides a comprehensive overview of the existing methodologies.Domain knowledge driven methods enhance LLMs' causal reasoning ability by leveraging domain-specific knowledge, such as input from domain experts, providing contextual knowledge, using pre-defined prompts, and fine-tuning LLMs. Model driven methods, on the other hand, improve LLMs' causal reasoning ability by integrating additional models, such as causal graph construction models, causal effect estimation models, and counterfactual reasoning models.
\\
In this section, we discuss these methods in detail. Furthermore, the strengths and weaknesses of each sub-category are summarized in Table \ref{tab:pros and cons}, while the annual publication trends for each class of methods are illustrated in Fig.\ref{fig:statistic_per_year}.

\begin{figure}[ht]
\centering
\includegraphics[width=1\linewidth]{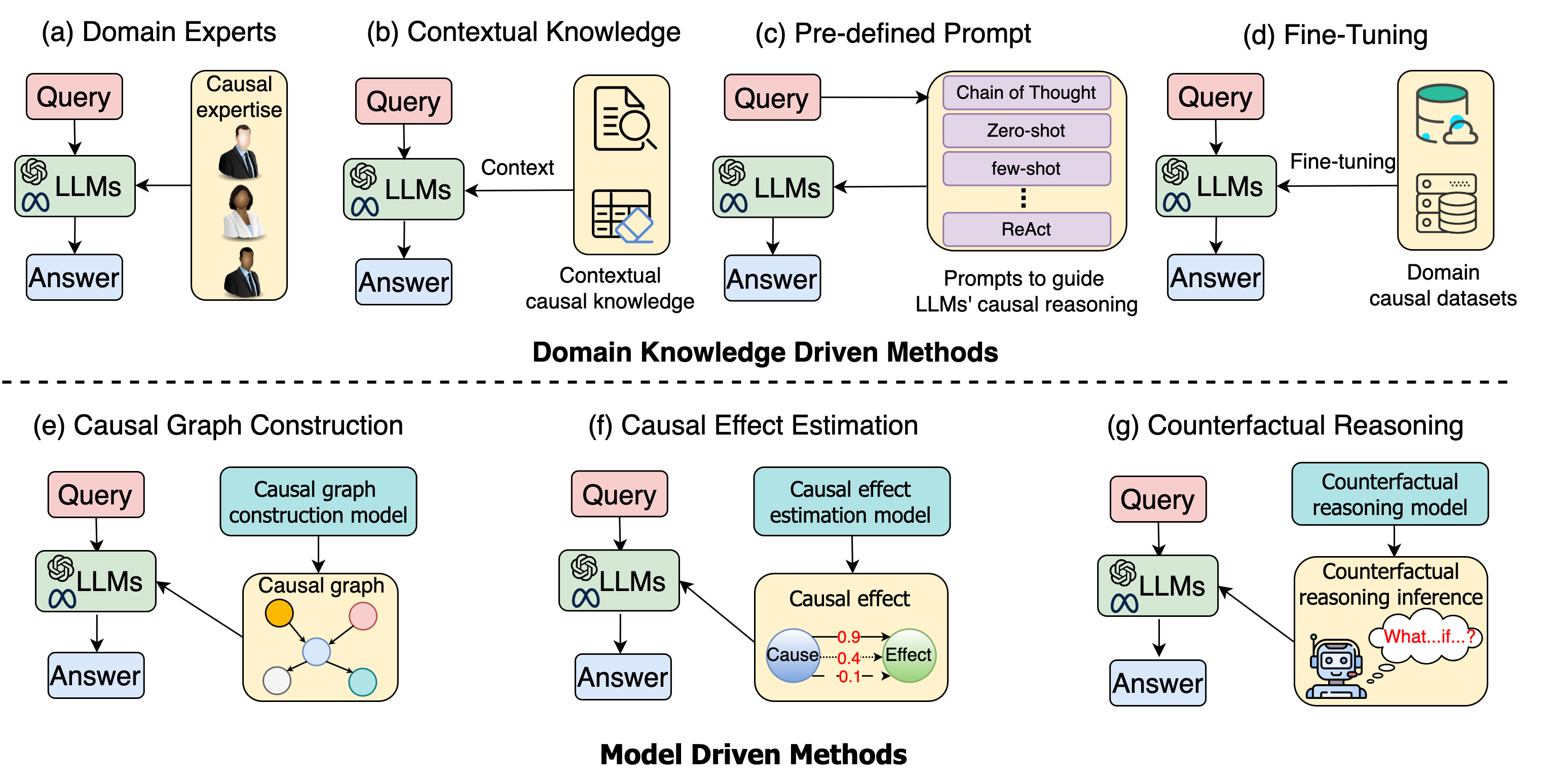}
\caption{Overiew of methods for enhancing LLMs’ causal reasoning ability}
\label{fig:overview}
\end{figure}

\subsection{Domain Knowledge Driven Methods}
Domain knowledge driven methods focuses on enhancing the causal reasoning ability of LLMs by incorporating real-time, domain-specific domain knowledge. This class of methods allows LLMs present robust causal reasoning ability that goes beyond the static knowledge in pre-trained datasets. According to different types of domain knowledge, this class of methods can be further divided into: domain experts, fine-tuning, contextual knowledge, and pre-defined prompts.

\subsubsection{Domain Experts}
\
\newline
The domain experts are professionals with deep knowledge and extensive practical experience in a specific field. Recently, a medication direction copilot (MEDIC) was developed based on the integration of LLMs and domain experts, utilizing strong reasoning ability to effectively prevent errors of medication direction \cite{pais2024large}. Besides, Castelnovo et al. \cite{castelnovo2023marrying} proposed a novel pipeline that incorporate a domain expert validation process to enhance LLMs' causal reasoning. 

Experts' knowledge can help LLMs identify and understand key causal relationships within datasets, particularly in complex or ambiguous situations. Through experts' feedback, LLMs can more accurately refine their causal inferences, avoiding wrong conclusions that may arise from a lack of background knowledge. Additionally, human intervention can help avoid ethical issues and ensure that the results generated by LLMs are aligned with human values. However, these methods require significant human and economic costs. Beyond this limitation, experts’ personal biases or subjective perspectives may compromise the accuracy of identifying causal relationships in LLMs. Furthermore, expert opinions often diverge, particularly in emerging fields, posing challenges to standardizing the verification process.

\subsubsection{Contextual Knowledge}
\
\newline
Contextual knowledge refers to the background information and understanding relevant to specific situations. This includes not only basic facts about the current topic but also insights into its history, culture, environment, social context, and other relevant factors. Yu et al. \cite{yu2024onsep} utilized the temporal causality from real-world events as contextual knowledge to enhance LLMs' causal reasoning, enabling them to effectively handle event prediction tasks. Furthermore, when analyzing temporal event sequences, LLMs can explore causality between events more effectively by incorporating pairwise qualitative causal knowledge as contextual knowledge \cite{shou2023pairwise}, the RealTCD framework can utilize domain knowledge as contextual support to enhance LLMs' ability to detect temporal causal relationships \cite{li2024realtcd}. Additionally, knowledge graph as injecting contextual knowledge to LLMs also can enhance LLMs' causal reasoning ability \cite{pan2024unifying,yu2024fusing}. 

Providing this broader contextual knowledge can increasingly enhance LLMs' performance in complex causal reasoning tasks. However, processing long texts with contextual knowledge significantly increase time and computational overhead. To address this issue, token compression offers a potential solution by optimizing models' computational efficiency or reducing costs through decreasing the number of tokens in the input or output text \cite{liu2023tcra,jiang2023longllmlingua}.

\subsubsection{Pre-defined Prompts}
\
\newline
To enhance LLM’s causal reasoning ability, pre-defined prompts play a crucial role in guiding LLMs to effectively identify causal relationships and perform causal reasoning tasks in specific domain \cite{cox2024ai}. Within the novel ALCM framework, a causal prompt is a text-based instruction designed to guide LLMs in generating specific outputs \cite{khatibi2024alcm}. Ban et al. \cite{ban2023query} developed a universal set of prompts with three stages, i.e., prompt understanding, prompt causal discovery, and prompt revision, to support LLMs in exploring causality from data. Additionally, Bagheri et al. \cite{bagheri2024c2p} proposed a causal chain of prompting (C2P) framework to strengthen LLMs's causal reasoning ability.

Designing proper prompts is a flexible and lightweight approach that can enhancing LLMs' causal reasoning ability without consuming extensive computational resources. Although effective, the effect of prompt engineering can be easily affected by subtle changes in prompts, leading to the inconsistency causal relationships generated by LLMs \cite{vashishtha2023causal,ji2023benchmarking}.

\subsubsection{Fine-tuning}
\
\newline
Fine-tuning involves using specific datasets to further train the pre-trained LLMs, and fine-tuned LLMs using specific datasets have shown increasing performance in causal reasoning tasks of target domains \cite{liu2024causal}, such as extracting causal relationships from medical guidelines \cite{gopalakrishnan2024causality}. Furthermore, integrating fine-tuning datasets with causal graph frameworks allows LLMs to achieve even better performance in causal reasoning tasks \cite{zheng2023preserving,bethany2024enhancing}. Additionally, based on the fine-tuning method, developing a novel LLM architecture can enhance robustness in handling complex out-of-distribution causal reasoning tasks \cite{gendron2024can}. 

Through this approach, LLMs can be adapted to perform causal reasoning tasks within particular fields more effectively. Despite remarkable success, fine-tuning LLMs requires significant computational resources. Besides, the training datasets for fine-tuning often need domain experts' knowledge to construct, which can be difficult and costly to obtain \cite{susanti2024prompt}.

\begin{table}[h]
    \centering
    \begin{tabular}{|c|p{2.8cm}|p{3.7cm}|p{3.6cm}|}
    \hline
        \multicolumn{2}{|c|}{Approach} & \multicolumn{1}{c|}{Strength}& \multicolumn{1}{c|}{Weakness} 
        \\ \hline
        \multirow{4}{*}[-55pt]{\makecell{Domain \\ Knowledge \\ Driven \\ Methods}}
        & Domain Experts - \cite{castelnovo2023marrying}, \cite{pais2024large}&Outputs align with human values and deliver domain-specific insights.& Hard to standardize the verification causal reasoning.
        \\ \cline{2-4}
        & Contextual Knowledge - \cite{yu2024onsep}, \cite{shou2023pairwise}, \cite{yu2024fusing},
        \cite{pan2024unifying}, \cite{li2024realtcd} & Generate informative responses and handle ambiguity.& Longer time to process contextual knowledge in causal reasoning process.
        \\ 
        \cline{2-4}
        & Pre-defined Prompt - \cite{bagheri2024c2p},\cite{ban2023query}, \cite{khatibi2024alcm}, \cite{cox2024ai}&Require less computational power and storage support.&Sensitive to prompt variations, leading to output inconsistencies.
        \\ \cline{2-4}
        & Fine-tuning - \cite{gopalakrishnan2024causality}, \cite{zheng2023preserving}, \cite{liu2024causal}, \cite{gendron2024can}, \cite{bethany2024enhancing}& Domain adaptation and bias reduction in LLMs.& Require significant computational resources and extensive datasets.
        \\ \cline{2-4}
        \hline
        \multirow{4}{*}[-40pt]{\makecell{Model\\Driven\\Methods}}
        & Causal Graph Construction - 
        \cite{jiralerspong2024efficient},
        \cite{darvariu2024large},
        \cite{ban2023causal},
        \cite{abdulaal2023causal}& 
        Effectively tackle complex causal reasoning with structured approaches.& Difficult to explore implicit causal relationships in real-world scenarios.
        \\
        \cline{2-4}
        & Causal Effect Estimation - \cite{chen2023ssl}, \cite{zhang2024causal}, \cite{liu2024discovery}, \cite{dhawan2024end} &  Predict intervention impacts to enhance result truthfulness.& Hard to be directly generalized to broader domains.
        \\ \cline{2-4}
        & Counterfactual Reasoning - \cite{ashwani2024cause}, \cite{cotta2024out} &Simulate potential outcomes of various interventions to improve reasoning accuracy.& Calculation of multiple hypothetical scenarios may lead to LLMs' performance degradation.\\ 
        \hline
    \end{tabular}
    \caption{A comparison of different classes of methods for enhancing LLMs' causal reasoning ability.}
    \label{tab:pros and cons}
    \vspace{-2em}
\end{table}

\subsection{Model Driven Methods}
Another research direction aims to enhance the causal reasoning ability of LLMs is based on causal learning models. This approach leverages causal structures and algorithms to develop novel frameworks to enhance LLMs' performance on causal reasoning tasks across diverse scenarios. Key methods explored for integrating causal models with LLMs include three primary classes: causal graph construction, causal effect estimation, and counterfactual reasoning.

\subsubsection{Causal Graph Construction}
\
\newline
A causal graph is a type of cause-and-effect diagram represented as a logic-based structure. The standard causal graph is typically a directed acyclic graph (DAG) without any loops, thereby preserving the transitivity and hierarchy of causal relationships. Recent researches show that LLMs have demonstrated the capability to construct causal graphs \cite{naik2024applying,zhang2024enhancing}. Moreover, Jiralerspong et al. \cite{jiralerspong2024efficient} proposed a novel method to construct causal graphs, leveraging a breadth-first search (BFS) approach integrated with LLMs.  Another approach integrates a probabilistic model of expert knowledge with LLMs to better construct causal graphs from observational datasets \cite{darvariu2024large}. Ban et al. \cite{ban2023causal} proposed a framework which utilizes an iterative process to combine LLMs' causal inference ability with causal structure learning, refining the causal graph through feedback from LLMs. Additionally, in the medical and healthcare fields, a causal modeling agent framework is constructed based on the integration of LLMs and deep structural causal models, which can significantly model the causal graph from the datasets of clinical and radiological phenotype of Alzheimer's Disease\cite{abdulaal2023causal}.

Causal graphs are instrumental in elucidating causal chains between variables and enhancing the causal reasoning ability of LLMs. However, many causal relationships are implicit and difficult to quantify explicitly. Due to the complexity of causal relationships in real-world, constructing accurate and complete causal graphs is very challenging and may lead to model bias or erroneous inferences. 

\subsubsection{Causal Effect Estimation}
\
\newline
The approach of causal effect estimation focuses on how much change in a "cause" leads to changes in an "effect". By calculating the causal effect, this class of approaches determine the outcome corresponding to a specific intervention. Zhang et al. \cite{zhang2024causal} proposed a novel method to calculate the causal effect between input prompts and output answers through the front-door adjustment to mitigate bias LLMs in causal reasoning tasks. 

Moreover, the causal effect estimation enables LLMs to more accurately explore relationships between variables, not only identifying the causality association but also estimating the strength of causal relationships. However, due to limitations in unstructured natural language data, such as network interference, outcome measurement and selection bias, estimated causal effects are often limited to specific groups and may not be directly generalizable to broader groups \cite{dhawan2024end}. To address this challenge, a causal representation assistant framework is introduced, which integrates LLMs with a causal discovery algorithm to improve causal effect estimation in unstructured datasets \cite{liu2024discovery}. Additionally, to address the issue regarding to the causal structure inconsistencies in unstructured data, a self-supervised learning (SSL) framework is developed to improve the accuracy in causal discovery tasks \cite{chen2023ssl}. 

\subsubsection{Counterfactual Reasoning}
\
\newline
The core concept of counterfactual reasoning involves making assumptions about observed outcomes and then reasoning to estimate the likelihood of one influencing factor. Counterfactual generation is essential in causality research, providing a key method for analyzing “what if” scenarios when LLMs perform causal reasoning \cite{li2023prompting}. Cotta et al. \cite{cotta2024out}integrated out-of-context prompting and counterfactual transformation methods to improve the trustworthiness of LLMs' generated outcomes in causal reasoning tasks. Additionally, Ashwani et al. \cite{ashwani2024cause} proposed a context-aware reasoning enhancement with counterfactual analysis (CARE-CA) framework, which is based on the integration of LLMs, explicit causal detection modules, contextual knowledge graphs and specific prompting mechanisms.

Through counterfactual reasoning models integrating, LLMs can evaluate the impact of changes in assumptions on outcomes. However, counterfactual reasoning often requires reasoning and calculation of multiple hypothetical scenarios, especially with large-scale observation datasets from complex scenarios, leading to performance degradation in causal reasoning tasks.

\section{Benchmarks and Evaluation Metrics}
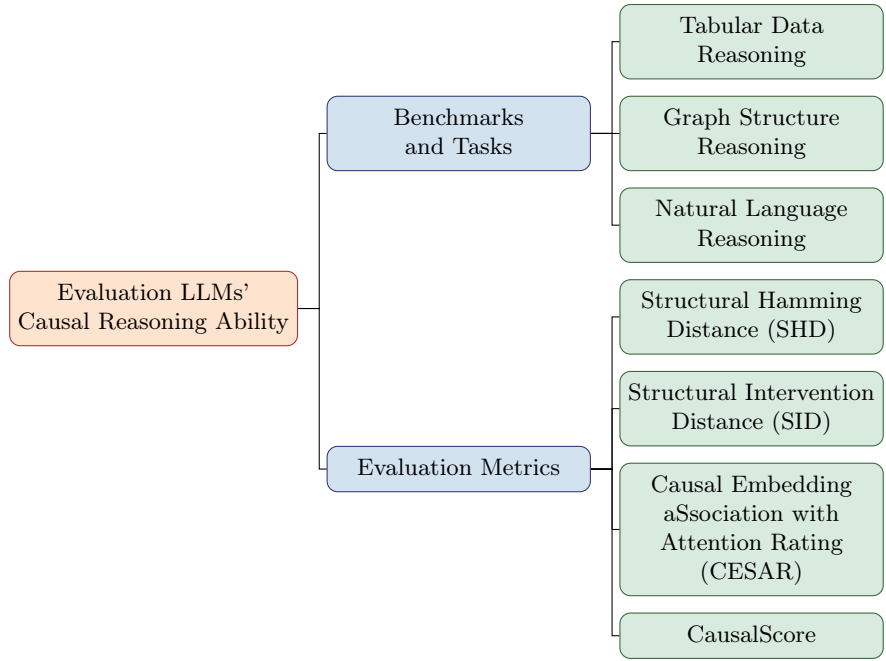
\begin{figure}
    \centering
    \begin{forest}
        for tree={grow=east, parent anchor=east, child anchor=west, draw, rounded corners, align=center,
        minimum width = 3.5cm,
        edge path={\noexpand\path [draw, \forestoption{edge}] (!u.parent anchor) -- ++(8 pt,0) |- (.child anchor)\forestoption{edge label};}
        }
        [Evaluation LLMs' \\Causal Reasoning Ability, draw=color1d, fill=color1
            [Evaluation Metrics,  draw=color2d, fill=color2            
                [CausalScore,  draw=color3d, fill=color3]
                [Causal Embedding \\aSsociation
                 with \\Attention Rating\\(CESAR),  draw=color3d, fill=color3]
                [Structural Intervention \\Distance (SID),  draw=color3d, fill=color3]
                [Structural Hamming \\Distance (SHD),  draw=color3d, fill=color3]
            ]
            [Benchmarks \\and Tasks,  draw=color2d, fill=color2
                [Natural Language \\Reasoning,  draw=color3d, fill=color3]
                [Graph Structure \\Reasoning,  draw=color3d, fill=color3]
                [Tabular Data \\Reasoning,  draw=color3d, fill=color3]
            ]
        ]
    \end{forest}
\caption{Classification of benchmarks and evaluation metrics for assessing LLMs' causal reasoning ability}
\label{fig:Evaluation}
\vspace{-2em}
\end{figure}

The evaluation of LLMs' causal reasoning ability plays an important role in advancing and standardizing the development of this area \cite{zhou2024causalbench}. In this section, we introduce the benchmarks and evaluation metrics for assessing LLMs' causal reasoning ability. The detail classification of different kinds of benchmarks and evaluation metrics in this area is shown in Fig.\ref{fig:Evaluation}.

\subsection{Benchmarks}
As shown in Fig.\ref{fig:Evaluation}, benchmarks for evaluating the causal reasoning ability of LLMs can be categorized into three main sub-categories: tabular data reasoning, graph structure reasoning, and natural language reasoning.
\subsubsection{Tabular Data Reasoning}
\
\newline
Real-world data patterns are typically stored as structured data in relational databases, organized in rows and columns. Consequently, evaluating the reasoning ability of LLMs with tabular data is crucial. To address this, the quantitative reasoning with data (QRDATA) benchmark was introduced to assess LLMs' causal reasoning ability in real-world statistical datasets. This benchmark includes a curated dataset of 411 questions derived from data sheets in textbooks, online learning resources, and academic articles \cite{liu2024llms}.
\subsubsection{Graph Structure Reasoning}
\
\newline
Causal graphs are a critical representation method for depicting causal relationships between variables. Furthermore, evaluating the extent to which LLMs understand causal graphs can provide insights into their causal reasoning ability. A novel benchmark dataset CLEAR \cite{chen2024clear}, has been introduced to facilitate this evaluation. It is categorized into three complexity levels and comprises 20 causal graph. Additionally, Jin et al. \cite{jin2023cladder} provided the CLADDER dataset which includes 10,000 samples based on causal graphs and related queries, offering a comprehensive assessment of LLMs' causal reasoning ability.

\subsubsection{Natural Language Reasoning}
\
\newline
Benchmarks and tasks based on natural language reasoning are designed to evaluate LLMs' causal reasoning ability. They typically adopt textual descriptions and question-answering formats. Chi et al. \cite{chiunveiling} proposed a new causal question-answering (QA) benchmark, CausalProbe-2024, to assess LLMs' human-like causal reasoning ability. Additionally, a novel CORR2CAUSE task has been introduced to evaluate LLMs’ causal reasoning skill about inferring causation from correlation within textual descriptions datasets\cite{jin2023can}.

\subsection{Evaluation Metrics}
Existing evaluation metrics that can be used to assess the causal reasoning ability of LLMs include structural hamming distance (SHD) \cite{zhou2024causalbench}, structural intervention distance (SID) \cite{zhou2024causalbench}, causal embedding aSsociation with attention rating (CESAR) \cite{cui2024exploring}, and CausalScore \cite{feng2024causalscore}, as shown in Fig.\ref{fig:Evaluation}.

\subsubsection{Structural Hamming Distance (SHD)}
\
\newline
The SHD metric quantifies the similarity between the predicted causal graph and the ground-truth graph by measuring the differences in their edges. This metric is particularly useful for evaluating the discrepancy between causal graphs generated by large language models (LLMs) and those derived from real-world data. The formula for calculating the SHD metric is as follows:
\begin{align}
\text{SHD}(G_1, G_2) = |E_1 \Delta E_2|,
\end{align}
where $G_1$ represents to the predicted causal graph and $G_2$ denotes the ground-truth graph structure. $E_1$ and $E_2$ are the edge sets of $G_1$ and $G_2$ respectively, and $\Delta$ denotes the symmetric difference. A lower SHD value indicates the causal graphs generated by LLMs are closer to the real causal relationships, which refers to stronger causal reasoning ability of LLMs.

\subsubsection{Structural Intervention Distance (SID)}
\
\newline
Compared to the SHD metric, the SID metric evaluates the similarity between causal graphs by counting inconsistencies in intervention distributions through various methodologies. The formula for calculating the SID metric is as follows:
\begin{align}
\text{SID}(G_1, G_2) = \# \left\{ (i, j) \mid i \neq j \right\},
\end{align}
where $G_1$ represents to the predicted causal graph and $G_2$ denotes the ground-truth graph structure. $(i, j)$ represents the pair of nodes between the generated causal graph by LLMs and the ground-truth causal graph. The SID metric quantifies how many interventions are required to transform the causal graph generated by LLMs into the ground-truth graph. Therefore, a lower SID value indicates stronger causal reasoning ability of LLMs.

\subsubsection{Causal Embedding aSsociation with Attention Rating (CESAR)}
\
\newline
The causal strength between variables, measured by the novel metric Causal Embedding Association with Attention Rating (CESAR), is determined through the weighted aggregation of token-level causal strengths. The formula for calculating the CESAR metric is as follows:
\begin{equation}
\begin{aligned}
\text{CESAR}(C \to E) = \sum_{i=1}^n \sum_{j=1}^ma_{ij}\{ c_i \to e_j \},
\end{aligned}
\end{equation}
where $C$ and $E$ represent the cause and effect tokenized sets, respectively. $c_i$ and $e_j$ are the corresponding elements of $C$ and $E$, and $a_{ij}$ denotes the attention weight assigned to each pair token of $c_i$ and $e_j$. The CESAR metric ranges from 0 to 1, where a larger value indicates the stronger intensity of causality between event $C$ and event $E$. Therefore, a larger CESAR value demonstrates stronger causal reasoning ability of LLMs.

\subsubsection{CausalScore}
\
\newline
CausalScore metric is introduced to assess the causal relevance between dialogue histories and responses within LLMs \cite{feng2024causalscore}.  The formula for calculating the CausalScore metric is as follows:
\begin{equation}
\begin{aligned}
\text{CausalScore}(Q, R) = \frac{1}{2} \Big( P_\text{Similarity}(Q_i, R_t) + P_\text{Similarity}(Q_i, Q_j, R_t) \Big),
\end{aligned}
\end{equation}
where $Q$ and $R$ represent the queries and responses, respectively, within the historical records of a dialogue.$P_\text{Similarity}(Q_i, R_t)$ represents the sum of dependencies between a single element from the $Q$ set and the $R$ set, and $P_\text{Similarity}(Q_i, Q_j, R_t)$ represents the sum of dependencies between pairwise elements from the $Q$ set and the $R$ set. 
A CausalScore value closer to 1 indicates a higher degree of causal relevance between the queries and responses, reflecting a stronger causal reasoning ability of the LLMs.

\section{Open Future Research Directions}
Existing methods have advanced the LLMs' causal reasoning ability to a great extent across various real-world scenarios. In this section, we further outline several promising research directions for future research.
\subsubsection{Improve Multi-modal Causal Reasoning Ability of LLMs}
\
\newline
Most existing LLMs primarily focus on causal reasoning within textual, tabular, or graph-based data structures. However, real-world data is predominantly multi-modal, encompassing text, images, videos, audio, and more. With advancements in LLMs, the focus of causal reasoning research is increasingly shifting toward multi-modal data. Multi-modal LLMs (MLLMs) have emerged as a prominent research area, as highlighted in studies like \cite{yin2023survey}. Additionally, the Multi-modal Causal Reasoning (MuCR) benchmark has been introduced to evaluate the causal reasoning capabilities of Vision Large Language Models (VLLMs), particularly when they rely exclusively on visual cues \cite{li2024multimodal}. Consequently, enhancing the causal reasoning ability of LLMs for multi-modal tasks represents a promising future research direction.

\subsubsection{Constructing Novel Memory Mechanisms for Improving LLMs' Causal Reasoning}
\
\newline
LLMs have demonstrated certain potential in causal reasoning and can be further strengthened through existing methodologies. However, they still face challenges in addressing long-range reasoning problems, primarily due to issues such as causal hallucinations and the vast search space involved \cite{wang2024credes}. A key contributing factor is the lack of a robust memory mechanism to effectively utilize external information. An innovative framework integrating a layered memory mechanism into LLMs has been proposed to support decision-making in financial domain \cite{li2023tradinggpt}. This study highlights the importance of constructing novel memory mechanisms to enhance LLMs' causal reasoning ability. Moreover, different causal reasoning tasks may require the development of specialized memory mechanisms to cater to their unique demands.

\subsubsection{Developing Novel Self-learning Mechanisms for LLMs}
\
\newline
Humans can learn and explore answers from unfamiliar learning materials when faced with unknown problems, demonstrating a form of self-learning ability, however, LLMs often rely on inherent knowledge derived from pre-trained datasets to tackle various tasks. As time progresses, LLMs' internal knowledge may become outdated, which can diminish their causal reasoning ability, particularly in specialized domains. Recent research suggests that the novel mechanism which is point in the unknown (PIU) can enhance  the self-learning ability of LLMs, which enhance their causal reasoning ability.\cite{ferdinan2024into}. Therefore, developing innovative self-learning mechanisms can emerge as a promising future research direction.

\subsubsection{Enhancing Ethical Alignment in LLMs' Causal Reasoning Processes}
\
\newline
With the advancement of LLMs, they are increasingly utilized in natural language understanding and generation of causal reasoning tasks. However, ensuring that the causal outcomes generated from natural language align with human values and ethical principles remains a significant challenge. For instance, a current study proposes a method for detecting toxic content in the outputs generated by pre-trained language models, utilizing logistic regression classifiers \cite{ousidhoum2021probing}.
Consequently, enhancing the ethical alignment of LLMs in causal reasoning tasks emerges as a promising future direction for future research.

\subsubsection{Developing Unified Evaluation Metrics and Training Datasets}
\
\newline
Research addressing different tasks often adopts diverse evaluation metrics to assess performance. However, this inconsistency in the choice of evaluation metrics complicates efforts to standardize research on enhancing LLMs' causal reasoning ability. Establishing unified metrics would facilitate more effective evaluations of LLMs' performance of causal reasoning. 

Additionally, the LLMs' causal reasoning ability is influenced by the quality and diversity of their training data. Constructing comprehensive training datasets spanning various domains allows LLMs to encounter a broad spectrum of causality scenarios during training, which can enhance their causal reasoning capabilities across diverse application contexts and overall generalization ability.

\section{Conclusion}
In recent years, various methods have been integrated with LLMs to enhance their causal reasoning ability, enabling them to tackle a range of downstream causal reasoning tasks effectively. This paper presents a comprehensive survey of existing approaches aimed at improving LLMs' causal reasoning ability. First, we discuss the motivations for enhancing these capabilities and highlight the necessity of addressing this area. Next, we identify key challenges involved in the enhancement process. Following this, we systematically summarize existing methods, categorizing them into domain-knowledge-driven and model-driven approaches, with each category further divided into specific subcategories. Additionally, we provide an overview of benchmarks and evaluation metrics used to assess the performance of these methods. Finally, we propose potential research directions to shed some light on the future research in this vibrant area.

%
%
%
\newpage
\bibliographystyle{splncs04}
\bibliography{ref}
%





\end{document}